%% file: verine22.tex
\documentclass[wcp]{jmlr}

\usepackage{longtable}

\usepackage{microtype}
\usepackage{booktabs} 
\usepackage{svg}
\usepackage{xr}
\usepackage{hyperref}
\usepackage{natbib}
\usepackage[english]{babel}
\usepackage{caption}

\makeatletter
\newcommand*{\addFileDependency}[1]{
  \typeout{(#1)}
  \@addtofilelist{#1}
  \IfFileExists{#1}{}{\typeout{No file #1.}}
}
\makeatother

\newcommand*{\myexternaldocument}[1]{%
    \externaldocument{#1}%
    \addFileDependency{#1.tex}%
    \addFileDependency{#1.aux}%
}

\myexternaldocument{verine22-supp}

\makeatletter
\let\Ginclude@graphics\@org@Ginclude@graphics 
\makeatother

\jmlrvolume{189}
\editors{Emtiyaz Khan and Mehmet G\"{o}nen}
\jmlryear{2022}
\jmlrworkshop{ACML 2022}

\title[On the expressivity of bi-Lipschitz normalizing flows]{On the expressivity of bi-Lipschitz normalizing flows}

  \author{\Name{Alexandre Verine} \Email{alexandre.verine@dauphine.psl.eu}\\
  \Name{Benjamin Negrevergne} \Email{benjamin.negrevergne@dauphine.psl.eu}\\
  \Name{Yann Chevaleyre} \Email{yann.chevaleyre@dauphine.psl.eu}\\
  \addr LAMSADE, Universit\'e Paris-Dauphine, PSL, Paris, France
  \AND
  \Name{Fabrice Rossi} \Email{rossi@ceremade.dauphine.fr}\\
  \addr CEREMADE, Universit\'e Paris-Dauphine, PSL, Paris, France
 }

\newcommand{\reals}{\mathbb{R}}
\newcommand{\R}{\mathbb{R}}

\newcommand{\Lip}{\mathrm{Lip}}
\newcommand{\inv}{^{-1}}
\newcommand{\Jac}{\mathrm{Jac}}
\newcommand{\KL}{\mathcal{D}_{\mathrm{KL}}}

\newcommand{\TV}{\mathcal{D}_{\mathrm{TV}}}

\newcommand{\Xset}{\mathcal{X}}

\newcommand{\ZZset}{\mathcal{Z}}

\newcommand{\vx}{\boldsymbol{x}}
\newcommand{\vzero}{\boldsymbol{0}}
\newcommand{\vz}{\boldsymbol{z}}
\newcommand{\vmu}{\boldsymbol{\mu}}
\newcommand{\vol}[1]{\mathrm{vol}(#1)}
\newcommand{\Supp}{\mathrm{Supp}}

\newcommand{\diag}[1]{\mbox{diag}(#1)}
\newcommand{\wh}{\widehat}

\begin{document}

\maketitle

\begin{abstract}
An invertible function is {\em bi-Lipschitz} if both the function and its inverse have bounded Lipschitz constants. Most state-of-the-art Normalizing Flows are bi-Lipschitz by design or by training to limit numerical errors (among other things). In this paper, we discuss the expressivity of bi-Lipschitz  Normalizing Flows and identify several target distributions that are difficult to approximate using such models. Then, we characterize the expressivity of bi-Lipschitz Normalizing Flows by giving  several lower bounds on the Total Variation distance between these particularly unfavorable distributions and their best possible approximation. Finally, we show how to use the bounds to adjust the training parameters, and discuss potential remedies.

\end{abstract}

\section{Introduction}
A weak smoothness condition for a function $F$, beyond continuity, can be enforced by requesting $F$ to be $L$-Lipschitz, that is to verify
\begin{equation*}
 \forall \vx_1,\vx_2 \in \Xset,\quad \Vert F(\vx_1) - F(\vx_2) \Vert_2 \leq L \Vert \vx_1 - \vx_2 \Vert_2,
\end{equation*}
where $L$ is the  Lipschitz constant of $F$.

A number of recent publications have demonstrated the benefits of constructing machine learning models with a small Lipschitz constant. First, models with a small Lipschitz constant have been linked with better generalization capabilities, both in terms of true risk \citep{bartlett_spectrally-normalized_2017}, and adversarial risk \citep{farnia_generalizable_2018}. In addition,  models with a small Lipschitz constants are more stable during training \citep{miyato_spectral_2018}, and are less prone to  numerical errors, a property which is particularly important in the context of invertible neural networks and normalizing flows \citep{behrmann_understanding_2021}.

Unfortunately, enforcing a small Lipschitz constant, either by design,
or using regularization during training, can impede the ability of
a model to fit the data distribution. Based on this observation,
several researchers have studied the limitations of neural networks
with bounded Lipschitz constants.  In particular, \citet{tanielian_learning_2020} were able to
identify a family of target distributions with disconnected support that cannot be fitted with a Generative Adversarial Networks (GANs) \citep{goodfellow_generative_2014} with a bounded Lipschitz constant. For  the particular case of normalizing flows, \citet{cornish_relaxing_2021} were able to demonstrate that there exist pairs of latent and target distributions with particular topological conditions on their support, that require the model to have an unbounded Lipschitz constant.

While the universality and the consistency of Normalizing Flow have been investigated \citep{kong_expressive_2020, zhang_approximation_2020, teshima_coupling-based_2020, koehler_representational_2021}, the numerical stability, i.e. the Lipschitz constraints are set aside.  Only \cite{kong_universal_2021} derive a clear link between a Lipschitz constant and the universality as a Maximum Mean Discrepancy of a subclass of Normalzing Flow. 

In this paper we focus on characterizing the impact of  Lipschitz constraints on the expressivity of normalizing flows. More precisely, we discuss the impact of the Lipschitz constant on the Total Variation distance (TV) between the approximated distribution and the target distribution. We give several  lower bounds on the TV distance which (unlike previous works), do not assume any hypothesis on the support of the target distribution, and are thus applicable to any learning settings.   Furthermore, since Normalizing Flows are often not only Lipschitz, but {\em bi-Lipschitz}, (meaning that both the inverse mapping function and the mapping itself have bounded Lipschitz constant), we also study the impact of the Lipschitz constant of the mapping, on the expressivity. (Most work on the topic focus on the Lipschitz constant of the {\em inverse} mapping.) Building on this analysis, we give an additional bound that depends on the Lipschitz constant of the mapping. 
Then, we use our bounds to exhibit practical limitations of Lipschitz constrained models on real world datasets such as {\em CIFAR10}, and we discuss the potential remedies in the light of these new results. Finally, we show how the bound can help choosing the depth and the Lipschitz constraint required for a particular dataset. 

To recap, our contribution is condensed to Theorems~\ref{BRL1cf}~and~\ref{BRL2}. We show that, with no specific hypothesis on the support, a normalizing flow will fail to perfectly approximate/generate if one the density conditions is met by the target distribution. 
The strength of the contribution is that we show that the Lipschitz continuity can also affect flows when the support are not disconnected. However, the limitation of the Theorems is that it can be hard in practice to check the density/sparsity properties of a high dimensional distribution.

\paragraph{Outline of the paper} The rest of this paper is organized as follows. Section~\ref{sec:background} reviews normalizing flows as well as total variation distance and the precision/recall for generative networks. The main results are presented in Section~\ref{sec:lowerbounds}: a first general bound is presented in the Section~\ref{subsec:boundA}, the two main theorems are presented and proved in the Section~\ref{subsec:boundbr}. We draw a link between the main results and the previous related work in the Section~\ref{subsec:relatedworks}. In Section~\ref{section:xp}, we make a expreimental analysis of the bounds. Finally we discuss the potential remedies to the highlighted limitations in Section~\ref{sec:discussion},  and Section~\ref{sec:conclusion} concludes the paper and gives some directions of future works. In the Supplementary Material, formal proofs of the different results are presented.




\section{Background}
\label{sec:background}
\subsection{Normalizing Flow}

A {\em  normalizing flow} is an invertible density model in which both density estimation and sampling can be done efficiently. In short,  training a normalizing flow consists in learning an invertible mapping between a data space $\Xset$ and a latent space $\ZZset$. Typically, the forward direction $F: \Xset \rightarrow \ZZset$ (i.e. the {\em normalizing} direction) is  tractable  and exact and the inverse direction $F^{-1}:\ZZset \rightarrow \Xset$ (i.e. the {\it generative} direction) either has a closed form, or can be approximated using an iterative algorithm. 

Suppose that $P^*$ is the true data distribution over $\Xset$, and that $P^*$ admits a density function denoted $p^*$ that we wish to approximate. We first choose a $d\mbox -$dimensional Gaussian distribution $Q$ over $\ZZset$ (a.k.a. the latent space), and its density function  $q(\vz) = \frac{1}{(\sqrt{2\pi})^d} e^{-\frac{1}{2}\Vert \vz \Vert^2_2}$. The choice of a normal distribution is natural since it is the most frequent latent distribution in normalizing flows. In Section {\ref{sec:discussion}}, more complex distributions will be discussed. Then,  we can define  $\hat p$, the  approximation  of $p^*$, based on $q$ and the mapping $F: \Xset \rightarrow \ZZset$, using a simple change of variable formula: 

\begin{equation}
\label{eq:cov}
\forall \vx \in \Xset, \quad \hat p(\vx) = \vert \det \Jac_F(\vx)\vert ~ q(F(\vx)).
\end{equation}

Note that the estimated probability $\wh P(A)$ of any event $A \subseteq \Xset$ can be computed as follows: 
$$
\widehat P(A) = Q(F(A)) = \int_{F(A)} q(\vz)d\vz.
$$

As seen in Equation~\ref{eq:cov}, performing density estimation requires computing the determinant of the Jacobian matrix which can be large in practice, thus most normalizing flows have been specifically designed to make this computation efficient.

\subsection{Bi-Lipschitz Normalizing Flows}
In this paper, we focus on bi-Lipschitz normalizing flows, which is a mapping $F$ whose Lipschitz constants are bounded in both directions. More specifically,  we define the bi-Lipschitz property as follows.
\begin{definition}
A bijective function $F:  \Xset \subset \reals^d  \rightarrow \mathcal{Z} \subset \reals^d $ is said to be $(L_1,L_2)$-bi-Lipschitz if $F$ is $L_1$-Lipschitz and its inverse $F\inv$ is $L_2$-Lipschitz, i.e.:
$$
\forall \vx_1,\vx_2 \in \Xset,\quad \Vert F(\vx_1) - F(\vx_2) \Vert_2 \leq L_1 \Vert \vx_1 - \vx_2 \Vert_2,
$$
\begin{center}
and  
\end{center}
$$
\forall \vz_1,\vz_2 \in \ZZset, \quad \Vert F\inv(\vz_1) - F\inv(\vz_2) \Vert_2 \leq L_2 \Vert \vz_1 - \vz_2 \Vert_2.
$$
\end{definition}

Alternatively, since the mapping $F$ is bijective, the bi-Lipschitz continuity can be expressed over $F$ only as: 
$$
\frac{1}{L_2}\Vert \vx_1 - \vx_2 \Vert_2 \leq \Vert F(\vx_1) - F(\vx_2) \Vert_2 \leq L_1 \Vert \vx_1 - \vx_2 \Vert_2.
$$
However, enforcing the bi-Lipschitz continuity of $F$ results in a bounded determinant for the Jacobian matrix:

\begin{proposition}
\label{prop:bounddet}
$\Jac_F$ satisfies for all $\vx \in \Xset\subset \reals^d$:
$$ \frac{1}{L_2^d} \leq \vert \mbox{\em det~} \Jac_F(\vx) \vert \leq L_1^d. $$
\end{proposition}
Proposition \ref{prop:bounddet} comes from a characterization of the Lipschitz continuity adapted to differentiable functions \citep{federer_geometric_1969}: 
$$\forall\vx \in \Xset, \,\Vert \Jac_F(\vx) \Vert_2 \leq L_1.$$ 
Therefore, through the spectral decomposition of the jacobian matrix, we can show that $\forall\vx \in \Xset$, $\vert \det \Jac_F(\vx) \vert \leq L_1^d $. Then, since $F$ is bi-Lipschitz, the same inequality can be expressed for $F\inv$: $\forall\vz \in \ZZset, \,\vert \det \Jac_{F\inv}(\vz) \vert \leq L_2^d$ and thus $\forall\vx \in \Xset$, $\vert \det \Jac_F(x) \vert \geq 1/L_2^d$.

As we will show in the rest of this paper, this can limit the expressivity of normalizing flows. 
This is relevant, because many normalizing flows are bi-Lipschitz in practice, for example, the i-ResNet  \citep{behrmann_invertible_2019} and the Residual Flow \citep{chen_residual_2020} are both based on residual atomic blocks $f_i = I_d + g_i$. Their invertibility is  ensured by the Lipschitz constant $\Lip(g_i) \leq L <1$. If $F$ is composed of $m$ residual blocks such that $F = f_m \circ \dots \circ f_1$, then the overall bi-Lipschitz constants satisfy $\Lip(F) \leq (1+L)^m$ and $\Lip(F\inv) \leq 1/(1-L)^m$. Alternatively, in Glow \citep{kingma_glow_2018} with atomic blocks $W_i = P_iL_i(U_i+\diag{s_i})$, the bi-Lipschitz constants satisfy: $\Lip(F) \leq \prod_i^m \Vert W_i \Vert_2$ and  $\Lip(F\inv) \leq \prod_i^m \Vert W_i\inv \Vert_2$. 

Notice that the bi-Lipschitzness constraints on either the function or its Jacobian determinant can frequently be weakened by increasing the depth of the network but, by doing so, the stability of the inverse can be affected \citep{behrmann_understanding_2021}.

\subsection{Assessing the expressivity}
Our goal is to understand how the bi-Lipschitz property affects the approximation ability of the network. To do so, we will compare the true data distribution $P^*$ and its density $p^*$ with the learned distribution  $\widehat P$ and its density $\hat p$.

In previous works, \cite{tanielian_learning_2020} use the maximum precision to evaluate how the true distribution $P^*$ and the generated distribution $\widehat P$ differs. 
We have chosen another tool to compare both distributions: the total variation (TV) distance given as~: 
\begin{definition}[Total Variation Distance]
\label{def:tv}
For any distribution $\widehat P$ and $P^*$,  the total variation distance is defined as the maximum difference of probabilities given to a same event $A$:
$$
\TV(P^*,\widehat P) =  \sup_A \vert P^*(A) - \widehat P(A) \vert. 
$$
\end{definition}
This choice has been made for several reasons. It is adequate to highlight the Lipschitz constraints of the mapping. It has a close connection with the precision and the recall, and yet, the TV is more general in terms of support of target distribution as explained in the followings. 
Finally, the TV can be used to compute a lower bound on the Kullback-Leibler divergence $\KL$ through the Pinsker's inequality: 
 $$
 2\TV(P^*, \wh P)^2 \leq \KL(P^*\Vert\wh P).
 $$

Even if the main results of this paper are consisting in lower bounds on the TV distance, we translate of result in terms of precision and recall. Therefore we provide their definitions as they were given initially \citep{sajjadi_assessing_2018, kynkaanniemi_improved_2019}.
\begin{definition}[Precision $\alpha$ and Recall $\beta$ for generative models]
For $\alpha, \beta \in [0, 1]$, the distributions $\widehat P$ is said to have a precision $\alpha$ at recall $\beta$ with respect to $P^*$ if there exist the distributions $\nu, \hat  \nu, \nu^*$, such that $\widehat P$ and $P^*$ can be decomposed as such:
$$
\widehat P = \alpha \nu + (1-\alpha)\hat  \nu \quad \mbox{and \quad }P^* = \beta \nu + (1-\beta)\nu^*.
$$

The distribution $\nu$ defined on $\Supp(\widehat P)\cup \Supp(P^*)$ while $\Supp(\hat  \nu) = \Supp(\widehat P) $ and $\Supp(P^*) =\Supp(\nu^*)$
\end{definition}
It can be interpreted as such: $\nu$ represent the part of $P^*$ that $\widehat P$ correctly models, $\hat  \nu$ is simultaneously the part of $P^*$ that $\widehat P$ misses on their joint support and all the points that should not be represented by $\widehat P$. Finally, $\nu^*$ cover the points of $P^*$ that the support of $\widehat P$ could not reach and all the points on their joint support that $\wh P$ misestimated. 

Among all the potential decompositions, i.e. the pairs $(\alpha, \beta)$, the focus is set on the maximum precision $\bar{\alpha}$ and the maximum recall $\bar{\beta}$.  

\begin{proposition}[Maximum precision $\bar{\alpha}$ and maximum recall $\bar{\beta}$]The maximum precision and the maximum recall satisfy:
$$
\bar{\alpha} = \widehat P (\Supp(P^*)) \quad \mbox{and} \quad \bar{\beta} = P^*(\Supp(\widehat P)).
$$
\end{proposition}

The results given by \cite{tanielian_learning_2020} is an upper bound on the maximum precision for a Lipschitz $F\inv$ and for a particular target distribution. Having upper bounds on $\bar \alpha$ or $\bar \beta$ is stronger than lower bounds on the $\TV$ since:
\begin{align*}
 \TV(P^*,\widehat P) \geq & \vert P^*(\Supp(P^*))  - \widehat P(\Supp(P^*))\vert,\\
 \TV(P^*,\widehat P) \geq & \vert P^*(\Supp(\widehat P)) - \widehat P(\Supp(\widehat P))\vert . 
\end{align*}
Or equivalently~: $\TV(P^*,\widehat P) \geq \max \left( 1-\bar{\alpha},1-\bar{\beta} \right)$

However, as soon as the support of, respectively,  the target distribution $P^*$ or the estimated distribution $\wh P$ covers $\Xset$, we have respectively $\bar \alpha = 1$ or $\bar \beta = 1$. Thus, the maximum precision/recall become irrelevant for assessing the expressivity of the normalizing flow. 

\section{Lower bounds on the TV distance}
\label{sec:lowerbounds}

The general idea is to look for subsets of the dataset in the data space $\Xset$ that may be particularly difficult to fit with a Lipschitz constrained mapping function. Intuitively, the
Lipschitz constraints limit the ability of normalizing flows to contract or to expand the latent space, so we focus our analysis on very dense subsets or very sparse subsets of the data space that will likely be the most difficult to fit. 

We focus first on dense subsets with arbitrary shape, and we are able to derive a positive lower bound on the TV that depends on the volume of the largest dense subset. Then, we show how to compute more specific but stronger results when considering subsets with a ball shape. First an intuitive but loose bound is derived. We then discuss two new tight bounds that are based on dense and sparse ball shaped subsets of the data space.
\subsection{A first bound based on the most dense subset of~$\Xset$}
\label{subsec:boundA}

The first theorem is a lower bound on the TV distance between the learned distribution and the target distribution in a general setting. Intuitively, the idea is to find a subset $A$ with an arbitrary shape that is sufficiently concentrated so that the Lipschitz constrained mapping can not concentrate enough weight from the Gaussian distribution onto this subset. 

\begin{theorem}[$L_1$-Lipschitz mappings fail to capture high density subset]
\label{AL1L2}
Let $ F$ be $L_1$-Lipschitz and $\eta_A = \frac{P^*(A)}{\vol A}$ be the average density over any subset $A \subset \R^d$. Then:
$$
\TV(P^*, \widehat P) \geq \sup_A \vol A \left( \eta_A - \left(\frac{L_1}{\sqrt{2\pi}}\right)^d\right).
$$
Therefore, if there is a subset $A$ that satisfies $\eta_A >\left(\frac{L_1}{\sqrt{2\pi}}\right)^d$, then the TV is necessarily strictly positive. 
\end{theorem}

Theorem \ref{AL1L2} results from the definition of the estimated distribution $\wh P$ and the change of variable. Indeed for an arbitrary subset $A$ of $\Xset$~:

$$\widehat P(A) =  \int_{A}\hat p(\vx)d\vx=   \int_{A}\vert \Jac_F(\vx)\vert q(F(\vx))d\vx.$$

Then, since $q$ is the density function of the normal distribution it is upper bounded by $1/\sqrt{2\pi}^d$, and with the upper bound of the determinant of the jacobian matrix given in Proposition~\ref{prop:bounddet}~:
$$
\wh P(A) \leq \left(\frac{L_1}{\sqrt{2\pi}}\right)^d \int_A d\vx =  \left(\frac{L_1}{\sqrt{2\pi}}\right)^d \vol A.
$$
In other terms, the weight assigned from the Gaussian latent distribution to the subset $A$ is bounded by $\left(L_1 /\sqrt{2\pi}\right)^d \vol A$. Consequently if there is a dense subset for which $P^*(A) = \eta_A \vol A$ is high enough, the  TV will be strictly positive. More precisely, the total variation is greater than the difference made by the most dense subset $A$. The detailed proof of Theorem~\ref{AL1L2} is given in Appendix~\ref{Proof:AL1L2}.

The main advantage of this formulation is to be applied to any subset of the data space, but at the expense of a loose bound on the TV.

\subsection{Bounds based on dense and sparse balls}
\label{subsec:boundbr}

The bound in Theorem \ref{AL1L2} can be further improved by making  assumptions on the structure of the subset $A$. We choose to focus on $l_2$ balls instead of arbitrary subsets. 

Let $B_{R, \vx_0}$ be the $l_2$ ball with center $\vx_0$ and radius $R$ (i.e. $B_{R, \vx_0} = \{\vx \in \Xset \,: \, \Vert \vx - \vx_0 \Vert_2 \leq R   \}$). Then we can show that both high density balls and low density ones are difficult to fit properly, the former because of the Lipschitz constraint of $F$, the latter because of the Lipschitz constraint of $F^{-1}$.

We first consider high density balls. 
\begin{theorem}[NF with a $L_1$-Lipschitz mapping $F$ fails to capture high density balls] 
\label{BRL1}
Let $F$ be $L_1$-Lipschitz. Then:
$$
\TV(P^*,\widehat P) \geq \sup_{R,\vx_0} \left(P^*(B_{R,\vx_0})- \frac{RL_1}{\sqrt{\pi}} \right). 
$$
Therefore, if we find a ball for which the true measure satisfies $\frac{P^*(B_{R,\vx_0})}{R}> \frac{L_1}{\sqrt{\pi}}$, then  the  TV  is  necessarily strictly positive.
\end{theorem}

\begin{figure}[t]
\centering
\includegraphics[width = \textwidth]{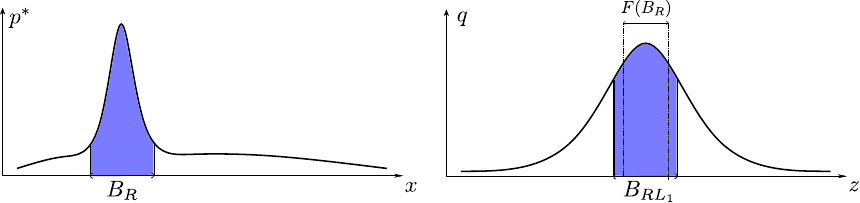}
\caption{Example of a target distribution where theorem \ref{BRL1} applies: the subset $B_R$ concentrates most of the weight in $P^*$, but $\wh P (B_R )=Q(F(B_R))$ can only be as large as $Q(B_{R{L_1}})$.}
\label{fig:BRL1}
\end{figure}

Theorem~\ref{BRL1} highlights the effect of the $L_1$ Lipschitz constraint of the forward mapping $F$. The image of a ball $B_R$ by the mapping $F$ is constrained in a ball:
$$
F(B_R) \subset B_{L_1R}.
$$
Thus, if we consider a ball with a high probability $P^*(B_R)$ in the data space, then the probability assigned to this ball  
$\hat{P}(B_R) = Q(F(B_R))$
is at most $Q(B_{RL_1})$ in the latent space and is upper bounded by $RL_1/\sqrt{\pi}$ \citep{ball_reverse_1993}. The lower bound of the TV in Theorem~\ref{BRL1} is given with a linear relation with radius $R$ and the Lipschitz constant $L_1$ in dimension $d$. It serves as an intuitive representation of the limitations induced by the Lipschitz constraint. For a given dense ball $B_{R,\vx_0}$, the smaller $R$ is, the greater $L_1$ should be to insure that the normalizing flow can properly map the Gaussian distribution onto $P^*$. The bounds mainly serve for interpretation purpose as we can compute a tighter bound on the TV distance. The closed form of Gaussian measure of a ball $B_{R,\vzero}$ is given by a function of the Gamma function $\Gamma$ and the incomplete gamma function $\gamma$, therefore:
$$
Q(B_{RL_1,\vx_0}) \leq Q(B_{RL_1,\vzero})=\gamma\left(\frac{d}{2},\frac{L_1^2R^2}{2}\right)/\Gamma\left(\frac{d}{2}\right).
$$
The Theorem~\ref{BRL1cf} is then less easy to interpret than Theorem~\ref{BRL1} but it proposes a tighter bound.
\begin{theorem}[NF with a $L_1$-Lipschitz mapping $F$ fails to capture high density balls] 
\label{BRL1cf}
Let $F$ be $L_1$-Lipschitz. Then:
$$
\TV(P^*,\widehat P) \geq \sup_{R,\vx_0} \left(P^*(B_{R,\vx_0})- \frac{\gamma\left(\frac{d}{2},\frac{L_1^2R^2}{2}\right)}{\Gamma\left(\frac{d}{2}\right)} \right).
$$

Therefore, if we find a ball for which the true measure satisfies $P^*(B_{R,\vx_0})> \gamma(\frac{d}{2},\frac{L_1^2R^2}{2})/\Gamma(\frac{d}{2})$, then the TV  is  necessarily strictly positive.
\end{theorem}

A one dimensional representation of a pathological case for Theorems~\ref{BRL1}~\&~\ref{BRL1cf} is shown on Figure~\ref{fig:BRL1}.  In other words no ball with a density high enough in the data space can be expanded sufficiently to have a matching probability in the latent space.

\begin{figure}[t]
\centering
\includegraphics[width =\textwidth]{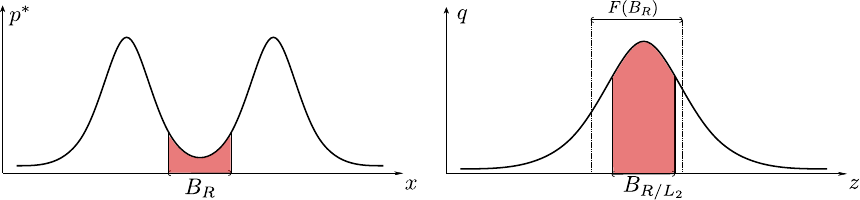}
\caption{Example of a target distribution for which Theorem~\ref{BRL2} applies: the subset $B_R$ concentrates little weight in $P^*$, but $\wh P (B_R) = Q(F(B_R))$ can only be as small as $Q(B_{R}/L_{2})$.}
\label{fig:BRL2}
\end{figure}

Conversely, the mapping being bi-Lipschitz, it can not contract arbitrarily:
$$
B_{R/L_2}\subset F(B_{R}).
$$
If there is a low density zone mapped on the maximum of the Gaussian density, then the Normalizing Flow cannot reduce enough the probability of the corresponding zone:
$$
\widehat P(B_R) \geq \widehat P( F\inv(B_R)) = Q(B_{R/L_2,0}).
$$
Notice that the assumption of a low density zone is strong but fairly reasonable. For instance, one can observe a multi-modal density with fairly well separated modes. If the modes are roughly equiprobable, we expect a mapping to assign those modes in balanced way around the mode of the Gaussian distribution in the latent space. Therefore, the low density ball is mapped on a zone wider than the ball $B_{R/L_2}$ and consequently the Gaussian measure associated is lower bounded by $Q(B_{R/L_2})$ as illustrated in the one dimensional example on Figure~\ref{fig:BRL2}. Despite the lower bounds established  by \citet{pinelis_exact_2020}, there is no reasonably interpretable bounds, therefore we use the closed-form as in Theorem~\ref{BRL1cf}. 

\begin{figure}
\begin{minipage}{0.45\textwidth}
\center
\includegraphics[width = 1\textwidth]{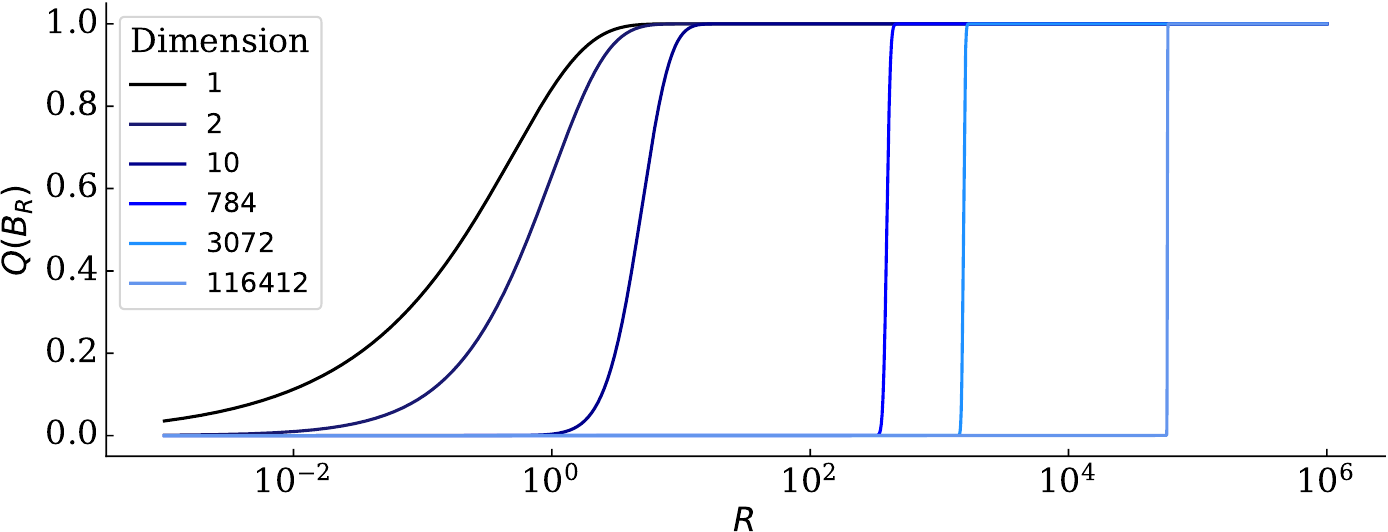}
	\caption{Representation of the Gaussian Measure of balls of radius $R$ centered on $\vzero$. The measure is given for dimension 1, 2, 10 and  then the dimensions of {\em MNIST} \citep{yann_lecun_mnist_2010}, {\em CIFAR10} \citep{alex_krizhevsky_learning_2009} and {\em CelebA} \citep{liu_deep_2015}}
\label{fig:gammainc}
\end{minipage}
\hfill
\begin{minipage}{0.45\textwidth}
\centering
\includegraphics[width =1\textwidth]{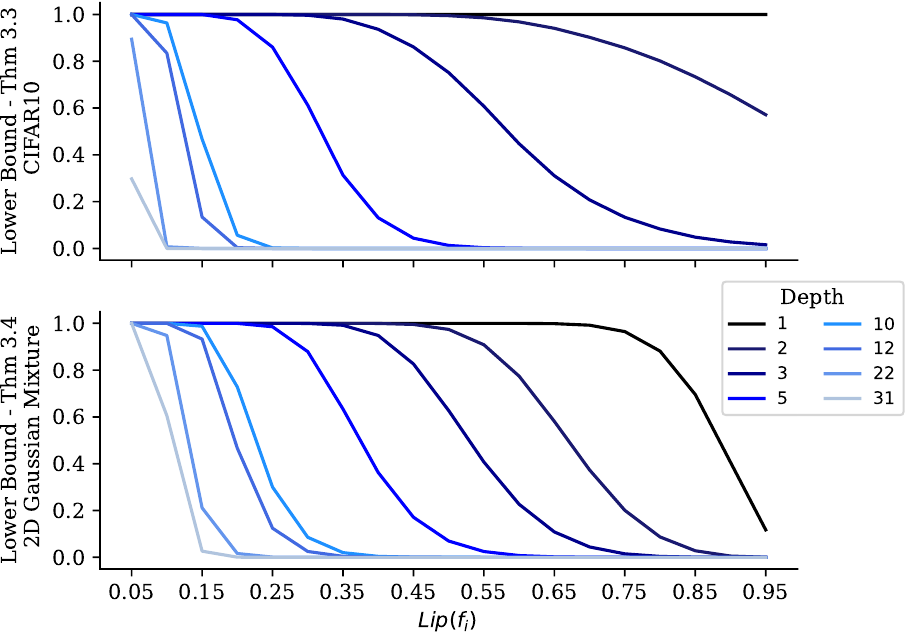}
\vspace{-1 \baselineskip}
	\caption{Lower bound of Theorem~\ref{BRL1cf} for {\em CIFAR10} (top) and Theorem~\ref{BRL2} for a 2D {\em Gaussian mixture} of Figure~\ref{fig:NGL} (bottom).}
\label{fig:BoundsL}
\end{minipage}

\end{figure}

\begin{theorem}[NF with $L_2$-Lipschitz inverse mappings $F\inv$  fail to capture low density balls] 
\label{BRL2}
Let $F\inv$ be $L_2$-Lipschitz. We consider the balls centered on $F\inv(0)$, we have the lower bound:
$$
\TV(P^*,\widehat P) \geq  \sup_{R}  \left( \frac{\gamma\left(\frac{d}{2}, \frac{R^2}{2L^2_2}\right)}{\Gamma\left(\frac{d}{2}\right)} - P^*(B_{R,F\inv(0)})\right).
$$ 
Therefore, if we find a ball for which the true measure satisfies $P^*(B_{R,F\inv(0)}) <  \frac{\gamma(d/2, R^2/2L^2_2)}{\Gamma(d/2)}$, then the TV is necessarily strictly positive. 
\end{theorem}
All formal proofs are detailed in Appendix~\ref{Proof:BL1}~\&~\ref{Proof:BL2}. Numerical illustrations of the behavior of the closed form bound are given in Figure~\ref{fig:gammainc}. Note that for Theorem~\ref{BRL1cf}, the probability $Q(B_{RL_1})$ needs to be as large as possible which, given a radius $RL_1$, will be harder while the dimension increases. In other terms, the Lipschitz constant $L_1$ will have to be consequently increased more in high dimension than in low dimension. The behavior is reversed for $L_2$: from Theorem~\ref{BRL2}, we see that the probability $Q(B_{R/L_2})$ should be as small as possible. Then, given a radius $R/L_2$, it will be harder in low dimension dimension than in high dimension. The Lipschitz constant $L_2$ will have a greater effect in low dimension than in high dimension as we will see in Section \ref{subsec:BRL2xp}

\subsection{Comparison to related work}
\label{subsec:relatedworks}

A related set up is used in the work of \citet{tanielian_learning_2020}. The authors consider two disconnected subsets $M_1$ and $M_2$ separated by a distance $D$, with equal probabilities in the latent space, i.e. $\widehat P(M_1) = \widehat P(M_1) = 1/2$. As a consequence, $F\inv(0)$ is equidistant from $M_1$ and $M_2$ as illustrated in Figure~\ref{fig:NGL}. The original work assesses the learning abilities of their generative model, a $L_2$-Lipschitz GAN \citep{goodfellow_generative_2014,arjovsky_wasserstein_2017}, with a definition of precision and recall \citep{sajjadi_assessing_2018, kynkaanniemi_improved_2019}. The authors propose an upper bound of the maximum recall based on the cumulative distribution function of the 1-dimensional normal distribution $\Phi(t) = \int_{-\infty}^t\frac{\exp(-r^2/2)}{2\pi}dr$:
$$
\bar{\alpha} +\frac{2D}{L_2}e^{-\Phi\inv(\bar{\alpha}/2)^2} \leq 1.
$$
Our method offers another bound on the maximum precision:
$$
\bar{\alpha} \leq 1 -  \frac{\gamma\left(\frac{d}{2},\frac{D^2}{2L_2^2}\right)}{\Gamma\left(\frac{d}{2}\right)}.
$$
The main advantage of our bound is that it can be directly computed whereas the bound given by \cite{tanielian_learning_2020} is not explicit.  The advantage of  their bound is that it does not depend on the dimension. The detailed proof can be found in Appendix~\ref{Proof:NGL}. From the link between the divergence and the maximum precision, we can derive a lower bound on the TV as a particular case of Theorem~\ref{BRL2}: 

\begin{figure}
\centering
\includegraphics[width =\textwidth]{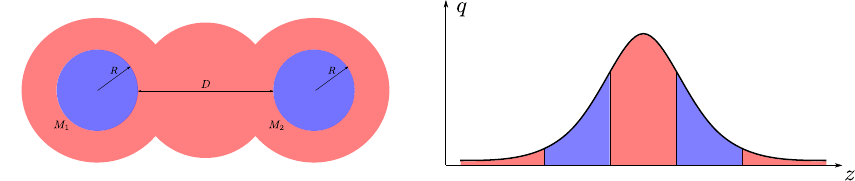}
\caption{Experimental set up given by \citet{tanielian_learning_2020}}
\label{fig:NGL}
\end{figure}
\begin{corollary}[NF with $L_2$-Lipschitz inverse mapping]
\label{NGL} If $F\inv$ is $L_2$-Lipschitz, then we have  a lower bound on the TV distance based on the distance $D$ between $M_1$ and $M_2$:
$$
\TV(P^*,\widehat P) \geq  \gamma\left(\frac{d}{2}, \frac{D^2}{2L^2_2}\right)/\Gamma\left(\frac{d}{2}\right).
$$
\end{corollary}

The main benefit of using maximum precision to assess the quality of the mapping is that it is well fitted to be used with the Gaussian Isoperimetric Inequality and therefore gives results that do not depend on the dimension $d$. The benefit of using TV distance is that we can now compute the bounds for distributions with  arbitrary support.
This covers more cases that the work by \cite{cornish_relaxing_2021} which only discusses the limits of the normalizing flows when the supports of the two distributions are {\em not} homeomorphic. Instead, the bounds we introduce in the present paper can be used to discuss the limitations of normalizing flows, even when dealing with two distributions with homeomorphic supports.
Let us consider for instance, the case where a normalizing flow tries to map a 1-dimensional Gaussian distribution with an excessively low or high variance onto the standard normal distribution: the supports are homeomorphic, the maximum precision and recall are both $1$, but our method can be used to derive a strictly positive lower bound of the TV.

%
%
\section{Experiments}
\label{section:xp}

Informally, 
our objective is to discover how deep and how constrained a network should be to fit a given dataset. To do so, we can compute the bounds for given theoretical Lipschitz constants. 
If the lower bounds are greater than zero, then the Lipschitz property will be limitating, and the settings should be adjusted. 


We focus on the Residual Flow \citep{chen_residual_2020}, for which the theoretical Lipschitz constant derivation is straightforward.  Two parameters will affect the two global Lipschitz constants $L_1$ and $L_2$ : the depth $m$ and the Lipschitz constant of the layers $L = \Lip(f_i)<1$ with $L_1\leq (1+L)^m$ and $L_2\leq1/(1-L)^m$. We compute the bounds for the two one-dimensional examples from Figure~\ref{fig:BRL1}~and~\ref{fig:BRL2}, for the pathological case from \citet{tanielian_learning_2020} in Figure~\ref{fig:NGL}, for the {\em 8 Gaussians} dataset and the {\em Circles} dataset shown in Figure~\ref{fig:2Dtoy} in Appendix and finally, for {\em MNIST} \citep{yann_lecun_mnist_2010} and {\em CIFAR10} \citep{alex_krizhevsky_learning_2009}.

\subsection{Bound of Theorem~\ref{BRL1cf}}
\begin{figure*}[h]
\centering
\includegraphics[width = \textwidth]{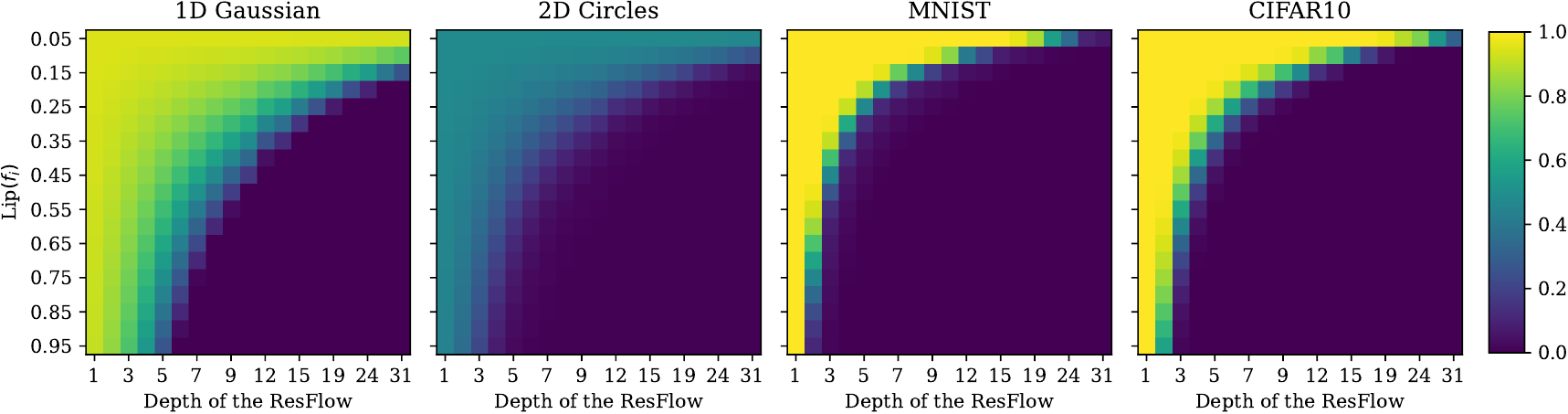}
\caption{Empirical values of the lower bound on the TV based on Theorem~\ref{BRL1cf}.}
\label{fig:BRL1emp}
\end{figure*}

The lower bound on the TV is based on the supremum over every ball in the dataset. To enumerate every candidate ball and find the supremum, we first consider every 
example $x$ in the dataset
as a candidate center and the distance to every other points in the dataset as a candidate radius $R$. Then we can compute an approximation of the true measure of the ball  $P^*(B_{R,x})$  with the empirical measure \citep{kloeckner_empirical_2018}. The results are presented in Figure~\ref{fig:BoundsL} and~\ref{fig:BRL1emp}.

In Figure~\ref{fig:BoundsL} (top), we can observe that if the network has a Lipschitz constant $L<0.5$ (usually required for stable training) it should be at least  10 layers deep. In contrast, a shallow network (1, 2 or 3 layers) with high Lipschitz constant will fail to capture the target distribution.

In Figure~\ref{fig:BRL1emp} we can see that for {\em 1D Gaussian},  the Residual Flow should be at least $7$ layers deep with an high atomic lipschitz constant of $0.95$. For {\em  2D Circles}, the inner circle is too dense to be properly mapped by the network. Since the inner circle represents half of the datapoints, the bound is at most $0.5$. Finally, we can see that the bound is limitating even on the real world datasets {\em MNIST} and {\em CIFAR10}. Moreover,
for shallow and highly constrained networks, the support of the learned distribution will not even intersect the support of the target distribution, as a consequence, TV (i.e. approximation error) will maximum. Instead,  when the depth or the atomic Lipschitz constant are increased, it results in a greater global constant and therefore in a reduction of the lower bound down to $0$.

\subsection{Bound of Theorem~\ref{BRL2}}
\label{subsec:BRL2xp}

The lower bound in Theorem~\ref{BRL2} relies on the supremum computed over all the balls centered on $F\inv(0)$, which can only be computed  after the normalizing flow has been trained. Thus the bound cannot be used to adjust the training parameters a priori as we did in the previous section. However,
the center of the probability mass is an intrinsic property of the dataset and we observe in practice that $F\inv(0)$ is often mapped onto the same point in the data space. For example, all the normalizing flows trained on the {\em 8~Gaussians} dataset will map the center of the Gaussian in the latent space onto the center of the 8 Gaussians in the data space. Building on this observation, we can compute $F\inv(0)$ analytically or train a first normalizing flow to find an estimate of $F\inv(0)$, and then compute the bound and adjust the parameters of the normalizing flow based on the instantiated bound.
For the one or two dimensional datasets, $F\inv(0)$ is clear, for {\em MNIST} and {\em CIFAR10}, $F\inv(0)$ is not obvious and can be seen in Figure~\ref{fig:f0MNIST}~and~\ref{fig:f0CIFAR} from the Appendix~\ref{sec:image}.

\begin{figure*}[h]
\centering
\includegraphics[width = \textwidth]{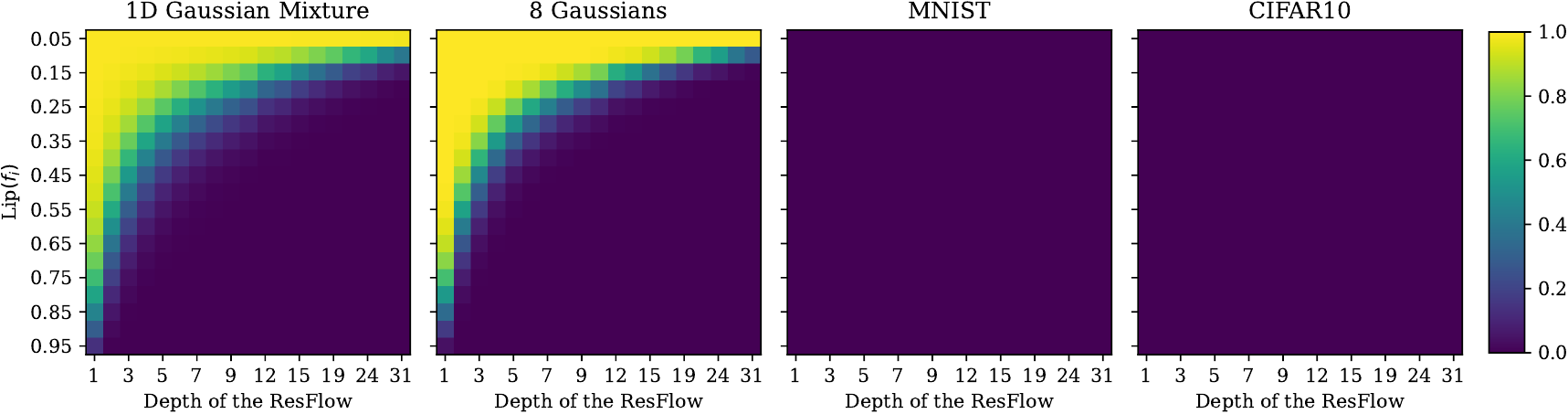}
\caption{Empirical values of the lower bound on the TV based on Theorem~\ref{BRL2}.}
\label{fig:BRL2emp}
\end{figure*}

The value of the bound computed for the different datasets are shown in Figure~\ref{fig:BoundsL} and \ref{fig:BRL2emp}. As we can see, this bound does not highlight any limitation for {\em MNIST} or {\em CIFAR10}, (as discussed in Section~\ref{subsec:boundbr}. However, as we will see in Section~\ref{sec:discussion},
one potential remedies introduces a tradeoff between the two bounds, and this bound may end up being limitating when it is used together with the other bounds. For the other two datasets, we can observe that for stable configuration with low $L$, the network needs to be at least 15 layers deep to learn the pathological case illustrated in Figure~\ref{fig:NGL}.


\section{Discussion on the potential remedies}
\label{sec:discussion}

As mentioned earlier, increasing the Lipschitz constants of the entire network (for example, by adding extra layers) may impact invertibility and stability during training \citep{behrmann_understanding_2021}, and thus is not a suitable approach to improve the expressivity. 

Alternatively, one can consider learning the parameters of the latent Gaussian distribution  $\vmu$ and $\Sigma = \sigma^2 I_d$. However, this is equivalent to changing the Lipschitz constants of $F$ from $(L_1,L_2)$ to $(\frac{L_1}{\sigma},L_2\sigma)$, thus this results in trading off the expected error on very dense subsets (Theorem~\ref{BRL1cf}) with the expected error on subsets with low densities (Theorems~\ref{BRL2}) or vice-versa. A proof of this statement is given in Appendix~\ref{Proof:leavar}. Increasing, reducing or learning the variance will indeed increase one bound and decrease the other one. In other words this can lead to a better approximation for some particular data distributions, but it does not generally improve the expressivity of the normalizing flow. 

To improve expressivity, a Gaussian Mixture latent distribution can also be considered. Indeed, \citet{khayatkhoei_disconnected_2019} and \citet{izmailov_semi-supervised_2019} have shown that such distributions can learn disconnected manifolds. When the latent distribution is a Gaussian Mixture, Theorem~\ref{BRL2} does not hold anymore.  Limitations similar to the ones highlighted in  Theorem~\ref{BRL1} still apply, but can be mitigated using learnable parameters.


We can trivially adapt the lower bound from Theorem~\ref{BRL1} to the Gaussian Mixture with  K equally distributed modes with learnable means $\vmu_i$ and covariance matrices $\Sigma_i = \sigma_{i}^2I_d$:

$$
\TV(P^*,\widehat P) \geq \sup_{R,\vx_0} \left(P^*(B_{R,\vx_0})- \frac{1}{K} \frac{\gamma\left(\frac{d}{2},\frac{ L_1^2R^2}{2\sigma_i^2}\right)}{\Gamma\left(\frac{d}{2}\right)} \right).
$$

As we can see here, the lower bound depends on the inverse of the number of modes $K$ in the mixture and the variance $\sigma_i$. Thus, this approach can solve the limitations highlighted in Theorem~\ref{BRL1cf}. The lower bound increases with $K$ but the learnable $\sigma_i$ can compensate this augmentation. However, learning a Gaussian Mixture suffers from the same issues than another method that we would like to mention: the Variational Mixture of Normalizing Flow \citep{pires_variational_2020}. They train set of $K$ different normalizing flows and a neural network with a K-class softmax output to set the mixture. They encountered strong training difficulties to learn the mixture and, as for the Gaussian Mixture, there is no efficient way yet to learn the hyperparameter $K$ \citep{izmailov_semi-supervised_2019}. A closely related method by \cite{dinh_rad_2020} consists in a discrete partition of the data space $\Xset$ into $K$ distinct subsets. $K$ different normalizing flows would be trained on each of those partitions. This methods also suffers from training difficulties and the lack of evidence on how to set the number of partitions. To avoid a discrete set of normalizing flows, \citep{cornish_relaxing_2021} propose a continuous set on normalizing flows, a promising method to tackle the limitations highlighted by Theorem~\ref{BRL2}, but it does not preserve an exact likelihood and requires a complex training process.

Overall, there exist some theoretical potential remedies to both of the limitations highlighted by  Theorems~\ref{BRL1cf}~\&~\ref{BRL2} but further investigation is required to to deal with the technical issues.

\section{Conclusion}

\label{sec:conclusion}
We have established that the bi-Lipschitz constraints reduce the expressivity of Normalizing flows. When the dataset meets some particular conditions such as a high density zone or a low density zone between two high density zones, the reduced expressivity fails to capture the real distribution of the dataset. More specifically, we have brought to light two particular lower bounds of the total variation distance between the target distribution and the approximated one. The first bound illustrates that dense subset, and especially dense balls in the data space can induce high approximation errors. The second bound illustrates that some low density balls located between high density subsets can also result in approximations errors. Finally, we showed that a more complex latent distribution,  such as a Gaussian Mixtures could solve the highlighted limitations. 

Moreover, Theorems~\ref{BRL1cf}~\&~\ref{BRL2} are based on the bi-Lipschitz constraints of the network and on general properties met by the dataset. Some similar bounds could be derived for non invertible structures that satisfy local bi-Lipschitzness. A possible direction for  future work is to generalize, not only the results, but also the framework to study the expressivity of generative lipschitz models based on the dataset.

Independently of the generative models, close attention is given to the effect of  Lipschitz regularization on  the stability of a neural networks \citep{scaman_lipschitz_2019,combettes_lipschitz_2020,bethune_many_2021}, on the  generalization capability of the network \citep{bartlett_spectrally-normalized_2017} or on its adversarial robustness \citep{szegedy_intriguing_2014, araujo_lipschitz_2020}. The theoretical framework presented in this work could be transposed to more general applications linked to Lipschitz regularization.

\bibliography{references}

\input{verine22-supp}
\end{document}

%% file: verine22-supp.tex
\title[Appendix: On the expressivity of bi-Lipschitz normalizing flows]{Appendix: On the expressivity of bi-Lipschitz normalizing flows}

\appendix
\renewcommand\thefigure{\thesection.\arabic{figure}}    
\section{Proofs}
\subsection{Proof of theorem \ref{AL1L2}}
\label{Proof:AL1L2}
By definition we have $\widehat P(A) =  \int_{A}\hat p(\vx)d\vx$, then with the change of variable formula we obtain~:

\begin{align*}
        \widehat P(A) & = \int_{A}\vert \Jac_F(\vx)\vert q(F(\vx))d\vx \\
     & = \frac{1}{(2\pi)^{d/2}}\int_{A}\vert \Jac_F(\vx)\vert e^{-\Vert F(\vx)\Vert^2/2}d\vx
\end{align*}

As $F$ is $L_1$-Lipschitz we have $\vert\Jac_F(\vx)\vert \leq L_1^d$, then
\begin{align*}
    \widehat P(A) &\leq \left(\frac{L_1}{\sqrt{2\pi}}\right)^d\int_{A}  e^{-\Vert F(\vx) \Vert_2^2}d\vx\\
     & \leq \left(\frac{L_1}{\sqrt{2\pi}}\right)^d \int_{A}d\vx \\
     & \leq \left(\frac{L_1}{\sqrt{2\pi}}\right)^d \vol A,
\end{align*}
and thus $TV(P^*, \widehat P)=\sup_A \vert P^*(A) - \widehat P(A) \vert$ implies
$$
TV(P^*, \widehat P) \geq  \\
\sup_A \left(P^*(A) - \left(\frac{L_1}{\sqrt{2\pi}}\right)^d \vol A \right) 
$$
\subsection{Proof of theorem \ref{BRL1}}
\label{Proof:BL1}
By definition of the TV distance, we have
$$
\TV(P^*,\widehat P) \geq  \sup_{R,\vx_0} \vert P^*(B_{R,\vx_0}) -  Q(F(B_{R,\vx_0})) \vert,
$$
where $B_{R,\vx_0}$ is the ball of a radius $R$ centered in $\vx_0$.

Then, the idea is to show that the image of a ball $B_R$ by a $L_1$-Lipschitz function is in a ball of radius $L_1R$, and then use a reverse isoperimetric inequality the find an upper bound of the measure of a ball of a radius $L_1R$.
\subsection*{Proof of $F(B_{R,\vx_0}) \subset B_{L_1R,F(\vx_0)}$  } First of all, for every $\vz\in F(B_{R,\vx_0})$, there exist $\vx\in B_R$ such that $F\inv(\vz) = \vx$, we have :
$$
\begin{array}{rcl}
    \Vert F(F\inv(\vz))- F(\vx_0) \Vert & =  &  \Vert F(\vx)- F(\vx_0) \Vert \\
     & \leq &   L_1 \Vert \vx - \vx_0 \Vert \\
     & \leq & L_1R
\end{array}
$$

\paragraph{Upper bound of $ Q(B_{L_1R})$} This bound is extracted from the work of \citet{ball_reverse_1993} on the Reverse Isoperimetric Inequality. First of all, it can be easily establish that $ Q(B_{L_1R}(F(\vx_0)))$ is at a maximum when $F(\vx_0)=0$. From now on,  we will only consider $B_{L_1R}$ the ball centered on $0$. Therefore the objective is to find an upper bound on :
$$
\begin{array}{rcl}
     Q(B_{L_1R}) & =  & \int_{\Vert \vz \Vert < {L_1R}}  q(\vz)d\vz \\
     & = & \int_{\Vert \vz \Vert < {L_1R}} \frac{1}{(\sqrt{2\pi})^d} e^{-\Vert \vz \Vert^2 /2}   d\vz  
\end{array}
$$
We can use the polar coordinates system to get another expression of the Gaussian measure with $S_{d-1}(r) = \frac{2\pi^{d/2}r^{d-1}}{ Q(d/2)}$ being the volume of the hypersphere : 
$$
\begin{array}{rcl}
    Q(B_{L_1R}) &= &\frac{1}{(2\pi)^{d/2}} \int_{0}^{L_1R} S_{d-1}(r)e^{-r^2/2} dr  \\
    & = & \frac{2}{2^{d/2} \Gamma(d/2)} \int_{0}^{L_1R} r^{d-1}e^{-r^2/2} dr
\end{array}
$$
However $r^{d-1}e^{-r^2/2}$ has a maximum value reached for $r=\sqrt{d-1}$, we can have an upper bound :
$$
\begin{array}{rcl}
    Q(B_{L_1R}) & \leq &   \frac{2}{2^{d/2} \Gamma(d/2)}\sqrt{d-1}^{d-1}e^{-\frac{d-1}{2}} \int_{0}^{L_1R}dr\\
     & \leq & \frac{\sqrt{2}{L_1R}}{ \Gamma(d/2)}  \left( \frac{d-1}{2e}\right)^{\frac{d-1}{2}}
\end{array}
$$
Then,  with the Stirling approximation of the Gamma function: 
\begin{align*}
    \frac{1}{2} \Gamma(d/2) & =  \frac{1}{d} \Gamma(d/2+1)  \\
     &\geq \frac{\sqrt{\pi} \sqrt{d}}{d}(d/2)^{d/2}e^{-d/2} \\
     & \geq  \frac{\sqrt{\pi}}{2^{d/2}}d^{\frac{d-1}{2}}e^{-\frac{d}{2}}
\end{align*}
We obtain:
$$
\begin{array}{rcl}
     Q(B_{L_1R}) &  \leq & \frac{2}{2^{d/2} \Gamma(d/2)}\left(d-1\right)^{\frac{d-1}{2}}e^{-\frac{d-1}{2}}  \\
     & \leq & \frac{{L_1R}\sqrt{e}}{\sqrt{\pi}}\left(\frac{d-1}{d}\right)^{\frac{d-1}{2}} 
\end{array}
$$
Using the bound
\begin{equation*}
\frac{1}{\sqrt{e}}< \left(\frac{d-1}{d}\right)^{\frac{d-1}{2}},   
\end{equation*}
we have
$$
 Q(B_{L_1R}) < \frac{L_1R}{\sqrt{\pi}}
$$

\paragraph{Lower Bound of the TV} As soon as we have an upper bound on $ Q(B_{L_1R})$, we have :
\begin{align*}
\TV(P^*,\widehat P) &\geq \sup_{R,\vx_0}\left(  P^*(B_{R,\vx_0}) - Q(F(B_{R,\vx_0})) \right)
     \\
     &\geq \sup_{R,\vx_0}\left(  P^*(B_{R,\vx_0}) - Q(B_{L_1R,\vx_0}) \right)
     \\
&\geq  \sup_{R,\vx_0}\left(  P^*(B_{R,\vx_0}) - \frac{L_1R}{\sqrt{\pi}} \right)
\end{align*}

\subsection{Proof of Theorem~\ref{BRL1cf}}
\paragraph{Value of $Q(B_{R,0})$}
By construction
\begin{equation*}
Q(B_{R,0})=\mathbb{P}\left(  \Vert \vz \Vert^2 \leq R^2\right),
\end{equation*}
when $\vz$ follows the standard Gaussian distribution in $\mathbb{R}^d$. This
quantity can be computed using the cumulative distribution function of the
chi-square distribution, i.e.
\begin{equation*}
Q(B_{R,0})=\frac{\gamma\left(\frac{d}{2},\frac{R^2}{2}\right)}{\Gamma\left(\frac{d}{2}\right)},
\end{equation*}
where $\gamma$ is the lower incomplete gamma function given by
\begin{equation*}
\gamma(x,k)=\int_{0}^xt^{k-1}e^{-t}dt.  
\end{equation*}

\paragraph{Lower Bound of the TV} Since we have the closed form of the measure over a ball we can write :
\begin{align*}
\TV(P^*,\widehat P) &\geq \sup_{R,\vx_0}\left(  P^*(B_{R,\vx_0}) - Q(F(B_{R,\vx_0})) \right)
     \\
     &\geq \sup_{R,\vx_0}\left(  P^*(B_{R,\vx_0}) - Q(B_{L_1R,\vx_0}) \right)
     \\
&\geq  \sup_{R,\vx_0}\left(  P^*(B_{R,\vx_0}) - \frac{\gamma\left(\frac{d}{2},\frac{L_1^2R^2}{2}\right)}{\Gamma\left(\frac{d}{2}\right)}\right)
\end{align*}
\subsection{Proof of theorem \ref{BRL2}}
\label{Proof:BL2}
In this section, we denote $B_R = B_{R,F\inv(0)}$. As $F\inv$ is $L_2$-Lipschitz, $F\inv(B_{R/L_2, 0})\subset B_R$ and thus
\begin{equation*}
\widehat P(B_R) \geq \widehat P( F\inv(B_R)) =  Q(B_{R/L_2,0}).
\end{equation*}
Therefore, by analogy with the proof of Theorem~\ref{BRL1cf}:
\begin{align*}
\TV(P^*,\widehat P) &\geq \sup_{R}\left( Q(F(B_{R})) -P^*(B_{R})\right) 
     \\
     &\geq \sup_{R}\left(Q(B_{R/L_2}) -  P^*(B_{R}) \right)
     \\
&\geq  \sup_{R}\left( \frac{\gamma\left(\frac{d}{2},\frac{R^2}{2L_2^2}\right)}{\Gamma\left(\frac{d}{2}\right)} - P^*(B_{R})\right)
\end{align*}\

\subsection{Proof of Corollary \ref{NGL}}
\label{Proof:NGL}
Since $M_1$ and $M_2$ are separated by a distance $D$ the ball centered on $ F\inv(0)$ has a radius at least as big as $D$ that we might call $B_D$ to simplify the notation. Therefore :
$$
\begin{array}{rcl}
    \bar{\alpha} & = & \widehat P(M_1) + \widehat P(M_2) \\
     & = & 1 - \widehat P(\overline{M_2\cup M_1}) \\
     & \leq & 1-\widehat P(B_D) \\
     &  \leq & 1 -  Q( F(B_D)) \\
     &  \leq & 1 -  Q(B_{D/L_2})) \\
     & \leq &  1 - \frac{\gamma(\frac{d}{2},\frac{D^2}{2L_2^2})}{\Gamma(\frac{d}{2})}
\end{array}
$$

And since $P^*(B_D) = 0$ :
$$
\begin{array}{rcl}
    \TV(P^*,\widehat P)  & \geq &    \vert \widehat P(B_D) - P^*(B_D) \vert \\
     & \geq &  \widehat P(B_D(F\inv(0)) \\
     & \geq & \frac{\gamma(\frac{d}{2},\frac{D^2}{2L_2^2})}{\Gamma(\frac{d}{2})}
\end{array}
$$

\subsection{Bounds for learned variance}
\label{Proof:leavar}
For a given variance $\sigma^2$ and the corresponding covariance matrix $\sigma^2I$, the Gaussian measure of a ball $Q_\sigma(B_R)$ of radius $R$ associated can be written as :
$$
\begin{array}{rcl}
     Q_\sigma(B_{R}) & =  & \int_{\Vert \vz \Vert < {L_1R}}  q_\sigma(\vz)d\vz \\
     & = & \int_{\Vert \vz \Vert < {R}} \frac{1}{(\sqrt{2\pi})^d\sigma^d} e^{-\Vert \vz \Vert^2 /2\sigma^2}   d\vz  
\end{array}
$$
Then, with the proper change of variable $z' = z/\sigma$, we have :
$$
\begin{array}{rcl}
     Q_\sigma(B_{R}) & = & \int_{\Vert\sigma\vz' \Vert < {R}} \frac{1}{(\sqrt{2\pi})^d\sigma^d} e^{-\Vert \vz \Vert^2 /2} \left\vert\sigma I \right\vert  d\vz'  \\
	& = & \int_{\Vert \vz \Vert < \frac{R}{\sigma}} \frac{1}{(\sqrt{2\pi})^d} e^{-\Vert \vz \Vert^2 /2}   d\vz  \\
	& = & Q(B_{R/\sigma})
\end{array}
$$

Hence the two bounds become :
$$
\TV(P^*,\widehat P) \geq \sup_{R,\vx_0} \left(P^*(B_{R,\vx_0})- \frac{\gamma(\frac{d}{2},\frac{ L_1^2R^2}{2\sigma^2})}{\Gamma(\frac{d}{2})} \right),
$$
and
$$
\TV(P^*,\widehat P) \geq  \sup_{R}  \left( \frac{\gamma(\frac{d}{2}, \frac{\sigma^2R^2}{2L^2_2})}{\Gamma(\frac{d}{2})} - P^*(B_{R,F\inv(0)})\right).
$$ 

\newpage
\section{2D datsets}
\begin{figure}[h]
\centering
\includegraphics[width =0.65\textwidth]{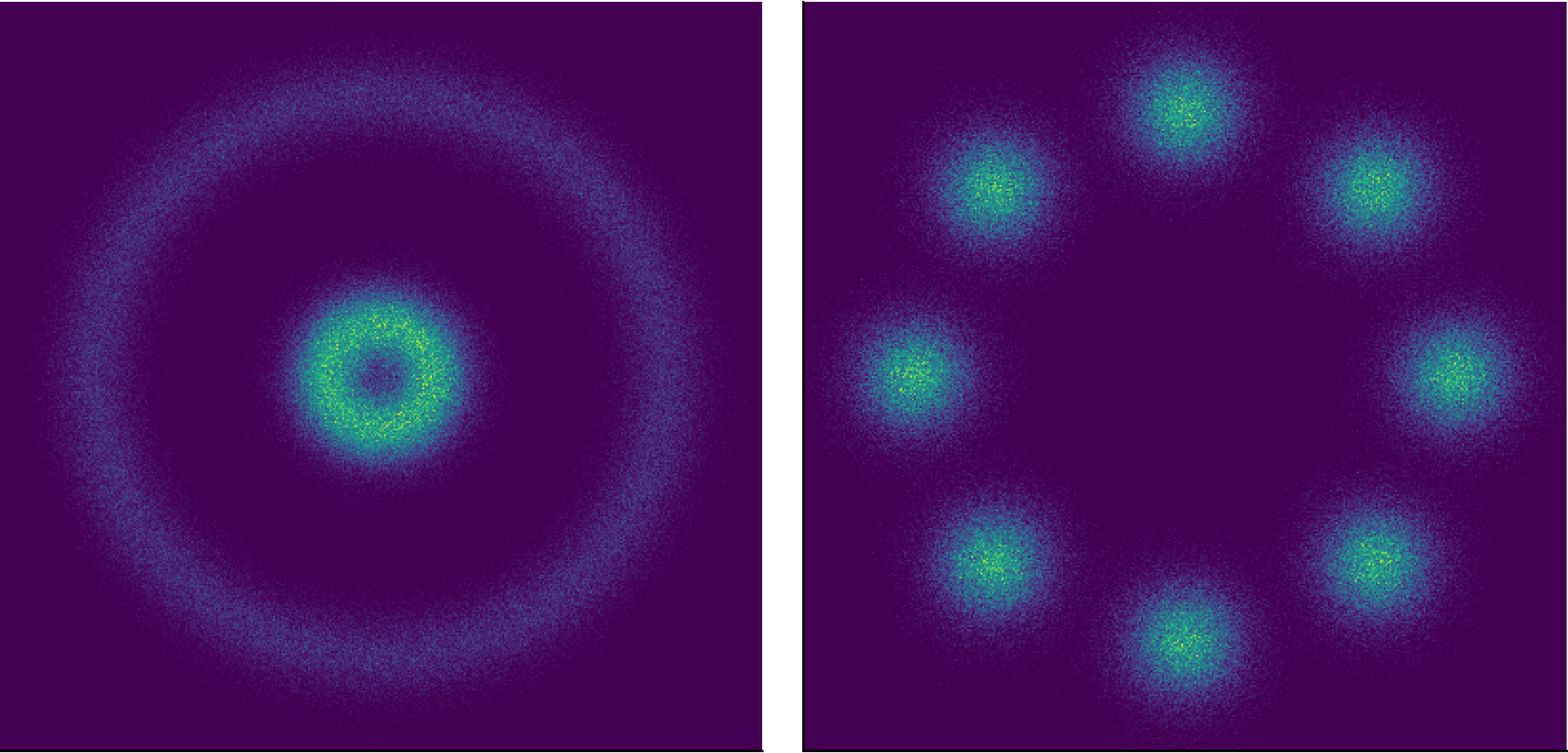}
	\caption{2D Dataset : {\em Circle}s (left) and  {\em 8 Gaussians} (right).}
	 \vspace{-1\baselineskip}
\label{fig:2Dtoy}
\end{figure}

\section{Inverse image of the center of the Gaussian latent distribution}
\label{sec:image}

\begin{figure}[h]
	\centering
	\includegraphics[width = 0.45\textwidth]{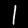}
	\caption{Image of $F\inv(0)$ for {\em MNIST} of the Residual Flow of \citet{chen_residual_2020} }
	\label{fig:f0MNIST}
\end{figure}

\begin{figure}[h]
	\centering
	\includegraphics[width = 0.45\textwidth]{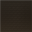}
	\caption{Image of $F\inv(0)$ for { \em CIFAR10} of the Residual Flow of \citet{chen_residual_2020} }
	\label{fig:f0CIFAR}
\end{figure}
